# GIAT: A Geologically-Informed Attention Transformer for Lithology Identification

*Abstract*—Accurate lithology identification from well logs is crucial for subsurface resource evaluation. Although Transformer-based models excel at sequence modeling, their "black-box" nature and lack of geological guidance limit their performance and trustworthiness. To overcome these limitations, this letter proposes the Geologically-Informed Attention Transformer (GIAT), a novel framework that deeply fuses data-driven geological priors with the Transformer's attention mechanism. The core of GIAT is a new attention-biasing mechanism. We repurpose Category-Wise Sequence Correlation (CSC) filters to generate a geologically-informed relational matrix, which is injected into the self-attention calculation to explicitly guide the model toward geologically coherent patterns. On two challenging datasets, GIAT achieves state-of-the-art performance with an accuracy of up to 95.4%, significantly outperforming existing models. More importantly, GIAT demonstrates exceptional interpretation faithfulness under input perturbations and generates geologically coherent predictions. Our work presents a new paradigm for building more accurate, reliable, and interpretable deep learning models for geoscience applications.

## I. INTRODUCTION

Lithology identification from well logs serves as a fundamental cornerstone for oil and gas exploration and reservoir evaluation, directly influencing hydrocarbon accumulation assessment and optimal drilling strategies [1]. Recent advances in deep learning, particularly Transformer architectures, have demonstrated remarkable potential for sequential data modeling, offering new paradigms for automated geological interpretation with enhanced accuracy and efficiency [2].

Recent advances in deep learning have introduced sophisticated Transformer-based architectures for lithology identification, achieving remarkable performance improvements. The Adaboost-Transformer method demonstrates superior accuracy rates of 95.20% and 95.50% compared to standalone models [3], while ReFormer integrates recurrent dynamics with multi-scale attention mechanisms for enhanced sequential data processing [4]. However, these approaches suffer from critical limitations that hinder their practical deployment in geological applications. First, they exhibit inherent "black-box" characteristics with unstable interpretation patterns, where attention mechanisms are fragile to adversarial perturbations and can be easily altered by slight input variations [5]. Second, current Transformer models lack geological domain constraints, operating purely on data-driven patterns without incorporating established geological principles. These limitations significantly impact geological decision-making processes, where interpretability and domain knowledge integration are essential for reliable subsurface characterization and risk assessment.

Parallel to deep learning approaches, geological prior-based feature engineering methods have demonstrated significant potential in lithology identification. The Category-wise Sequence Correlation (CSC) filter represents a data-driven, model-free approach that captures sequence structural variations by leveraging labeling information to enhance discrimination between different lithology types [6]. Domain knowledge (DK) constraint methods integrate geological expertise through formation lithology indices (FLI) to filter and optimize training data, achieving improved accuracy and stability in well log reconstruction [7]. Additionally, correlation analysis combined with multi-scale median filtering techniques effectively extract geological features while reducing noise interference, enabling BiLSTM networks to better capture bidirectional sequence dependencies [8]. Wavelet transform-based approaches further enhance model interpretability by incorporating stratigraphic information, with studies showing average performance improvements of 6.25% through geological information integration[9]. However, these geological prior methods face critical limitations in practical applications. They exhibit insufficient integration with deep learning architectures, often operating as preprocessing steps rather than being deeply embedded within model structures. Furthermore, their utilization of temporal and contextual information remains inadequate, failing to fully exploit the sequential nature and spatial correlations inherent in well logging data, ultimately constraining their performance potential.

Analysis of current methodologies demonstrates that both deep learning and geological prior-based approaches possess distinct advantages and limitations in lithology identification. Deep learning models effectively capture complex nonlinear patterns and achieve high prediction accuracy. For example, attention-based convolutional neural networks can prioritize key features in seismic data [10], but their interpretability is

Jie Li is with the College of Petroleum, China University of Petroleum-Beijing at Karamay, Karamay 834000, China (e-mail: 2142411064@qq.com).
Qishun Yang is with the College of Petroleum, China University of Petroleum-Beijing at Karamay, Karamay 834000, China (e-mail: 2131742581@qq.com).

Nuo Li is with the College of Petroleum, China University of Petroleum-Beijing at Karamay, Karamay 834000, China (e-mail: 692099794@qq.com).

limited and they lack geological domain constraints. In contrast, geological prior methods enhance model interpretability and domain knowledge integration, and have improved accuracy in metamorphic rock classification tasks [11]. However, these methods are insufficiently integrated with deep learning architectures and do not fully exploit temporal and spatial dependencies. Efficient integration of geological priors with deep learning models, balancing performance and interpretability, has become a central scientific task in lithology identification. Most current fusion attempts remain at the preprocessing stage and lack deep architectural integration. Therefore, a unified framework that organically combines the pattern recognition capabilities of deep learning with the domain expertise embedded in geological priors is essential for advancing lithology identification technology.

To address the critical scientific challenge of efficiently integrating geological prior knowledge with deep learning models, this letter presents the Geologically-Informed Attention Transformer (GIAT) framework. The core innovation lies in transforming data-driven geological priors from CSC filters into explicit attention bias matrices that are directly injected into the Transformer's self-attention mechanism, achieving deep architectural integration rather than superficial preprocessing. By constructing a geological similarity matrix and transforming it into a dynamic bias matrix M, GIAT fundamentally regularizes the learning process, addressing the interpretation instability issues identified in existing Transformer architectures while ensuring that model predictions are both data-driven and constrained by geological principles[12]. Comprehensive experiments on two challenging datasets demonstrate GIAT's superior performance in both accuracy and interpretation faithfulness, achieving 94.7% accuracy on the Kansas dataset (surpassing the best baseline by 3.9%) and exceptional interpretation stability under input perturbations, with PCC improvements of 46.6% and 62.9% over standard Transformer models, fundamentally resolving the trade-off between performance and interpretability that has long plagued geological deep learning applications.

## II. Methodology

As illustrated in Fig. 1, GIAT integrates three core components. Module (a) learns category-specific CSC filters from well logs to establish geological prior templates. Module (b) generates geological attention maps based on these filters and combines them with standard Transformer attention to create a guidance matrix $M$. Module (c) utilizes context vectors refined through this geologically-guided attention to produce final lithology predictions. The model's core innovation lies in this synergy: offline-captured prior knowledge is transformed online into explicit attention bias, thereby regularizing the learning process. This ensures that final predictions are both data-driven and constrained by geological principles, enhancing accuracy and interpretability.

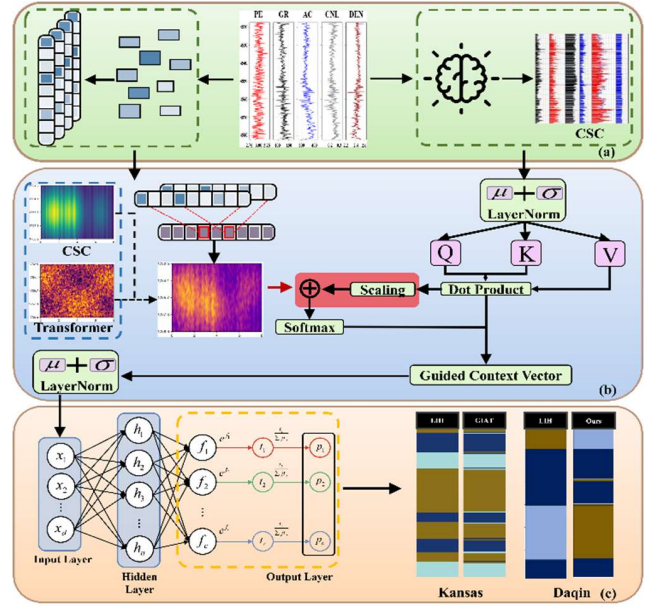

**Fig. 1.** The overall architecture of GIAT. (a) CSC Filter Learning; (b) Geological Attention Fusion; (c) Geologically-Informed Self-Attention

### A. CSC Filter Learning

We use CSC filter to establish geological priors. This data-driven tool captures unique sequential structural features for each lithology class from training data statistics. Each filter represents a typical pattern learned for a specific class on a specific curve. Unlike learned convolutional kernels, these filters provide a stable and interpretable prior. Rather than direct feature extraction, we use them as expert templates to evaluate geological similarity across the sequence.

### B. Geological Attention Fusion

The core of our method converts prior knowledge into a dynamic attention bias matrix $M$, addressing interpretation instability in existing Transformer architectures [13]. The input sequence is convolved with pre-computed CSC filters to generate a response map tensor. A geological feature vector $g(u)$ is constructed for each sequence position $u$ by concatenating response values, quantifying the match between local data patterns and geological prototypes. A relational similarity matrix $S \in \mathbb{R}^{L \times L}$ is constructed by computing cosine similarity between feature vectors:

$$S_{i,k} = \frac{g(i) \cdot g(k)}{\| g(i) \| \; \| g(k) \|} \tag{1}$$

where $S_{i,k}$ represents the geological consistency between positions $i$ and $k$. Finally, this similarity matrix $S$ is processed to form the final attention bias matrix $M$.

*C. Geological Attention Fusion*

Standard self-attention mechanisms are data-agnostic. Our GIAT model addresses this limitation by directly incorporating the bias matrix $M$ into the attention score calculation, as shown in the following modified formula:

$$\text{Attention}(Q,K,V) = \text{softmax}\left(\frac{QK^T}{\sqrt{d_k}} + M\right)V \quad (2)$$

By adding matrix $M$ before softmax normalization, we provide a strong inductive bias that guides the model to focus on geologically similar position pairs. High values in $M_{i,k}$ enhance the corresponding attention scores $(QK^T)_{i,k}$, explicitly directing attention toward positions with similar geological characteristics. This fundamentally reshapes the learning process, where model attention is no longer learned from scratch but actively guided by geological principles. The mechanism ensures that learned attention patterns are both data-driven and geologically meaningful, improving model performance and interpretive fidelity.

*D. Model Training Objective*

GIAT is trained end-to-end as a multi-class classification task. The final layer of the model is a fully connected layer with a softmax activation function, which outputs the probability distribution over the $C$ lithology classes. The model's parameters are optimized by minimizing the standard cross-entropy loss $\mathcal{L}$, a widely used objective function for classification problems, defined as:

$$\mathcal{L} = -\sum_{i=1}^{N}\sum_{c=1}^{C} y_{i,c} \log(\hat{y}_{i,c}) \quad (3)$$

where $N$ is the total number of samples, $C$ is the number of classes, $y_{i,c}$ is a binary indicator of the true class, and $\hat{y}_{i,c}$ is the model's predicted probability for that class.

III. EXPERIMENTS AND RESULT

This section details the experimental framework, presents the quantitative and qualitative results, and analyzes the performance of our proposed GIAT model.

*A. Experimental Setup*

To ensure comprehensive and rigorous evaluation, our experiments are conducted on two distinct types of datasets: the public Kansas cross-well dataset and a more challenging private dataset from heterogeneous reservoirs in the Daqing Oilfield. For both datasets, we adopt a strict cross-well validation strategy, reserving one well as a completely independent blind test well to evaluate the model's generalization capability.

We benchmark our proposed GIAT against baseline methods. Model performance is primarily evaluated through standard classification metrics, including accuracy, precision, recall, and Kappa. Additionally, to quantitatively assess interpretation faithfulness, we design a perturbation experiment. When introducing small bounded noise to the input, we measure the stability of model attention maps by computing Pearson Correlation Coefficient (PCC) and Structural Similarity Index (SSIM). All deep learning models are implemented in PyTorch and trained using the Adam optimizer with a learning rate of 1e-4, employing early stopping mechanisms to prevent overfitting.

*B. Results and Analysis*

(1) Quantitative Comparison

As presented in Table I, GIAT exhibits significant superiority over baseline methods. On the Kansas dataset, GIAT achieved an accuracy of 94.7%, surpassing DRSN-GAF by 3.9%. On the Daqing Oilfield dataset, it attained an accuracy of 95.4% and a Kappa coefficient of 0.94—an improvement of 6.5% over DRSN-GAF. This demonstrates its robustness and generalization capability in complex heterogeneous reservoir conditions. GIAT outperforms both ReFormer and ResGAT approaches, underscoring the importance of integrating geological prior information.

TABLE I
Performance comparison on two datasets

| Model | Dataset | Accuracy | Precision | Recall | Kappa |
|---|---|---|---|---|---|
| BiLSTM | Kansas | 79.3 | 78.1 | 78.9 | 0.73 |
| | Daqing | 76.8 | 75.6 | 76.4 | 0.70 |
| ResGAT | Kansas | 81.7 | 80.5 | 81.3 | 0.75 |
| | Daqing | 79.1 | 77.9 | 78.7 | 0.72 |
| Transformer | Kansas | 83.8 | 82.6 | 83.4 | 0.78 |
| | Daqing | 81.5 | 80.3 | 81.1 | 0.75 |
| ReFormer | Kansas | 86.2 | 85.0 | 85.8 | 0.81 |
| | Daqing | 84.1 | 82.9 | 83.7 | 0.78 |
| DRSN-GAF | Kansas | 90.8 | 89.6 | 90.4 | 0.87 |
| | Daqing | 88.9 | 87.7 | 88.5 | 0.85 |
| GIAT | Kansas | 94.7 | 93.5 | 94.3 | 0.92 |
| | Daqing | 95.4 | 94.2 | 95.0 | 0.94 |

(2) Interpretation Faithfulness

Fidelity validation results are presented in Table II. On the Kansas dataset, GIAT achieved PCC of 0.85 and SSIM of 0.82, improving by 19.7% and 7.9% over ReFormer. On the Daqing oilfield dataset, GIAT attained PCC and SSIM of 0.88 and 0.85, enhancing by 31.3% and 18.1% over ReFormer. DRSN-GAF's complex structure generated larger activation changes under perturbations, while Transformer-based models lacked geological constraints, resulting in poor fidelity. These findings demonstrate that geological prior knowledge with fidelity constraints improves classification accuracy and regularizes attention, making interpretations robust for geological decision-making.

TABLE II
Attention faithfulness comparison under input perturbation

| Model | Dataset | PCC | SSIM |
| --- | --- | --- | --- |
| DRSN-GAF | Kansas | 0.39 | 0.46 |
|  | Daqing | 0.36 | 0.43 |
| Transformer | Kansas | 0.58 | 0.36 |
|  | Daqing | 0.54 | 0.59 |
| ReFormer | Kansas | 0.71 | 0.76 |
|  | Daqing | 0.67 | 0.72 |
| GIAT | Kansas | 0.85 | 0.82 |
|  | Daqing | 0.88 | 0.85 |

(3) Qualitative Visualization

Fig. 2 presents lithology prediction comparison for several deep learning models on a challenging interval from the Kansas oilfield dataset. Panels (a)-(e) show the five input well logs (PE, GR, AC, CNL, DEN). Our proposed GIAT (m) achieves superior prediction highly consistent with ground truth lithology (f). In contrast, other models exhibit varying degrees of error. DRSN-GAF (k) incorrectly identifies sandstone layers within mudstone formation around 4565m. This problem is exacerbated in Transformer (i) and BiLSTM (g), which produce fragmented and geologically discontinuous predictions. This visualization underscores our model's architectural superiority in integrating geological context for coherent lithological interpretations.

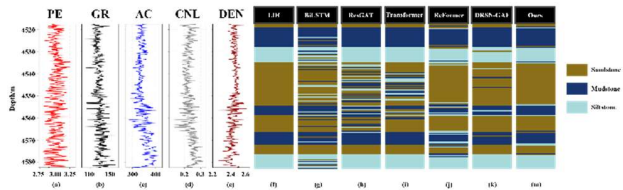

**Fig. 2.** Lithology prediction comparison on Kansas oilfield dataset. (a)-(e) Input well logs (PE, GR, AC, CNL, DEN); (f) Ground truth lithology; (g)-(m) Model predictions from BiLSTM, ResGAT, ReFormer, Transformer, DRSN-GAF, and GIAT respectively.

To validate interpretation faithfulness, we conducted perturbation experiments as shown in Fig. 3. Panels (b)-(e) show original predictions, while (f)-(k) display results after adding Gaussian noise. DRSN-GAF, Transformer, and ReFormer exhibit poor faithfulness, producing fragmented artifacts under perturbation due to overfitting to noise. Our model maintains structural coherence and geological consistency with ground truth (a), demonstrating superior stability through geologically-informed architecture.

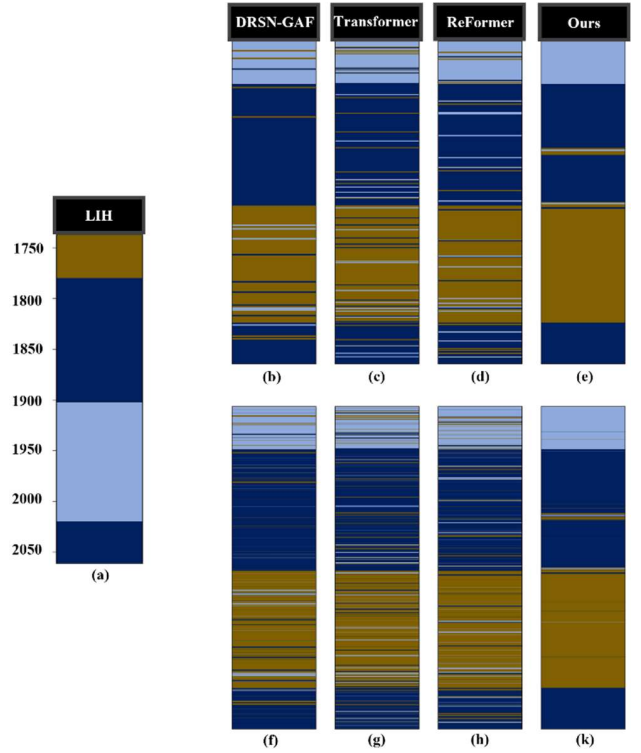

**Fig. 3.** Model robustness comparison under input perturbation. (a) Ground truth lithology; (b)-(e) Original predictions from DRSN-GAF, Transformer, ReFormer, and GIAT; (f)-(k) Corresponding predictions after Gaussian noise perturbation. GIAT demonstrates superior stability with minimal prediction changes under perturbation.

IV. ANALYSIS OF RESULTS

This section analyzes the experimental results, evaluating the proposed GIAT against baselines in terms of classification performance, component efficacy via ablation studies, and advancements in interpretability and faithfulness.

*A. Quantitative Performance Analysis*

GIAT's superior performance in Table I stems from its novel architecture. Unlike BiLSTM that suffers from information bottlenecks, GIAT employs global attention for comprehensive contextual understanding across geological sequences. GIAT overcomes standard Transformer limitations by explicitly injecting geologically-guided attention bias, addressing their instability tendency as unconstrained learning struggles to capture geological patterns. The bias matrix stabilizes learning and ensures focus aligns with subsurface structural continuity. This prior knowledge integration is more direct than geologically-blind transformations in DRSN-GAF and ResGAT.

*B. Ablation Study Analysis*

The ablation studies confirm the critical role of geology-guided attention bias. Removing bias matrix M led to accuracy drops of 10.9% and 13.9% on Kansas and Daqing datasets, respectively. This underscores data-agnostic attention limitations. The matrix imposes geological constraints rather than simply boosting performance. Without constraints, attention lacks geological context and becomes susceptible to spurious correlations. Removing bias degraded accuracy and reduced interpretability, causing PCC values to decrease by over 30%. This simultaneous decline indicates bias matrix M fundamentally regularizes attention, ensuring high performance is grounded in geological principles.

*C. Interpretation Faithfulness and Qualitative Analysis*

The primary contribution is achieving both high accuracy and trustworthy interpretability. As shown in Table II, GIAT's interpretations remain exceptionally stable under input perturbation, improving PCC by 62.9% over Transformer. This quantitative robustness is visually corroborated in Figure 2, where baseline models degrade into fragmented, geologically implausible patterns, while GIAT preserves coherent geological layering consistent with the ground truth. The superior faithfulness is a direct outcome of the architectural design, where M constrains the attention mechanism to prioritize robust geological relationships over spurious noise. This ensures the resulting interpretations are stable and reliable for practical applications.

V. CONCLUSION

This work presents GIAT, a novel framework addressing fundamental challenges in deep learning-based lithology identification—lack of domain knowledge constraints in Transformer architectures and shallow application of geological knowledge extraction. Our core innovation transforms data-driven geological priors from CSC filters into explicit attention bias matrices, injected into Transformer's self-attention mechanism to guide toward geologically coherent patterns. Experimental validation demonstrates GIAT's advantages over baseline models. On Kansas dataset, GIAT achieves 94.7% accuracy, surpassing DRSN-GAF by 3.9%. On Daqing Oilfield dataset, GIAT's accuracy exceeds DRSN-GAF by 6.5%. GIAT achieves breakthroughs in interpretation faithfulness, with PCC values improving 46.6% and 62.9% over Transformer. Ablation studies confirm that removing geological bias matrix causes simultaneous degradation in accuracy and faithfulness, demonstrating our approach fundamentally regularizes attention mechanism. While GIAT represents significant progress, limitations exist. Model performance depends on training data quality for CSC filters. The framework requires adaptation for other geological interpretation tasks. Future work will explore diverse geological priors integration and model transferability across geological settings.